\documentclass[lettersize,journal]{IEEEtran}
\usepackage{amsmath,amsfonts}
\usepackage{algorithmic}
\usepackage{array}
\usepackage[caption=false,font=normalsize,labelfont=sf,textfont=sf]{subfig}
\usepackage{textcomp}
\usepackage{stfloats}
\usepackage{url}
\usepackage{verbatim}
\usepackage{graphicx}

\usepackage{hyperref}
\usepackage{booktabs}
\usepackage{multirow}
\usepackage{bbding}
\usepackage{subfig}
\usepackage{adjustbox}
\usepackage{caption}
\usepackage{tabularx}
\usepackage{academicons}

\hypersetup{
    colorlinks=true, 
    linkcolor=blue,  
    citecolor=blue,  
    urlcolor=blue    
}
\makeatletter
\renewcommand{\@cite}[2]{\textcolor{blue}{[{#1\if@tempswa, #2\fi}]}}
\makeatother

\usepackage{xcolor,colortbl}
\definecolor{myred}{RGB}{249,217,218}
\definecolor{myblue}{RGB}{218,232,252}
\usepackage{listings}
\newlength\savewidth\newcommand\shline{\noalign{\global\savewidth\arrayrulewidth
  \global\arrayrulewidth 1pt}\hline\noalign{\global\arrayrulewidth\savewidth}}
\usepackage{paralist}
\input{def.set}

\usepackage{xspace}

\makeatletter
\DeclareRobustCommand\onedot{\futurelet\@let@token\@onedot}
\def\@onedot{\ifx\@let@token.\else.\null\fi\xspace}

\def\eg{\emph{e.g}\onedot} 
\def\ie{\emph{i.e}\onedot}

\makeatother

\def\BibTeX{{\rm B\kern-.05em{\sc i\kern-.025em b}\kern-.08em
    T\kern-.1667em\lower.7ex\hbox{E}\kern-.125emX}}
\usepackage{balance}
\begin{document}
\title{VideoXum: Cross-modal Visual and Textural Summarization of Videos}
\author{Jingyang Lin \and Hang Hua \and Ming Chen \and Yikang Li \and Jenhao Hsiao \and Chiuman Ho \and Jiebo Luo
}
\author{
 Jingyang Lin$^{\href{https://orcid.org/0009-0000-3223-3827}{\includegraphics[scale=0.06]{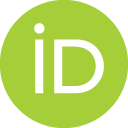}}}$, {\it Graduate Student Member, IEEE}, Hang Hua$^{\href{https://orcid.org/0000-0002-5441-5776}{\includegraphics[scale=0.06]{images/icon/ORCID_icon.png}}}$,  Ming Chen$^{\href{https://orcid.org/0000-0003-2603-5239}{\includegraphics[scale=0.06]{images/icon/ORCID_icon.png}}}$,
 Yikang Li$^{\href{https://orcid.org/0000-0003-4666-9642}{\includegraphics[scale=0.06]{images/icon/ORCID_icon.png}}}$ \\ \vspace{-0.5mm}
 Jenhao Hsiao$^{\href{https://orcid.org/0009-0003-4542-7809}{\includegraphics[scale=0.06]{images/icon/ORCID_icon.png}}}$,
 Chiuman Ho$^{\href{https://orcid.org/0000-0002-1607-6080}{\includegraphics[scale=0.06]{images/icon/ORCID_icon.png}}}$, Jiebo Luo$^{\href{https://orcid.org/0000-0002-4516-9729}{\includegraphics[scale=0.06]{images/icon/ORCID_icon.png}}}$, {\it Fellow, IEEE} \\
 \vspace{-6mm}
\thanks{\scriptsize Manuscript received 31 July 2023; revised 8 October 2023; accepted 12 November 2023; data of current version 21 March 2024. The Associate Editor coordinating the review of this manuscript and approving it for publication was Wu Liu. {\em(Jingyang Lin and Hang Hua contributed equally to this work.)} {\em(Corresponding author: Jiebo Luo.)}

J. Lin, H. Hua, and J. Luo are with the Department of Computer Science, University of Rochester, Rochester, NY 14627 USA (e-mail: jlin81@ur.rochester.edu; hhua2@ur.rochester.edu; jluo@cs.rochester.edu).

M. Chen, Y. Kang, J. Hsiao, and C. Ho are with OPPO US Research Center, Palo Alto, CA 94303 USA (e-mail: cmelf0819@gmail.com; lyk010632@gmail.com; mhsiao.pro@gmail.com; chiuman100@gmail.com).

The project page is \url{https://videoxum.github.io/}.

Digital Object Identifier 10.1109/TMM.2023.3335875}
}

\markboth{IEEE TRANSACTIONS ON MULTIMEDIA, VOL. 26, 2024}%
{How to Use the IEEEtran \LaTeX \ Templates}

\maketitle

\begin{abstract}
Video summarization aims to distill the most important information from a source video into either an abridged video clip or a textual narrative.
Existing methods often treat the generation of video and text summaries as independent tasks, thus neglecting the semantic correlation between visual and textual summarization.
In other words, these methods only study a single modality as output without considering coherent video and text as outputs.
In this work, we first introduce a novel task: cross-modal video summarization. This task seeks to transfer a long video into a condensed video clip and a semantically aligned textual summary, collectively referred to as a cross-modal summary. 
We then establish VideoXum (X refers to different modalities), a new large-scale human-annotated video benchmark for cross-modal video summarization. VideoXum is reannotated based on ActivityNet Captions with diverse open-domain videos.
In the current version, VideoXum provides 14K long videos, with a total of 140K pairs of aligned video and text summaries. Compared to existing datasets, VideoXum offers superior scalability while preserving a comparable level of annotation quality.
To validate the dataset's quality, we provide a comprehensive analysis of VideoXum, comparing it with existing datasets.
Further, we perform an extensive empirical evaluation of several state-of-the-art methods on this dataset. Our findings highlight the impressive generalization capability of the vision-language encoder-decoder framework yields on VideoXum. Particularly, we propose VTSUM-BLIP, an end-to-end framework, serving as a strong baseline for this novel benchmark. Moreover, we adapt CLIPScore for VideoXum to measure the semantic consistency of cross-modal summaries effectively.
\end{abstract}


\section{Introduction}
\label{sec:intro}

\IEEEPARstart{V}{ideo} summarization, which is known as generating a concise summary that conveys the primary parts of a full-length video, is a profound challenge for video analysis. Practical automatic video summarization systems have a great potential impact on numerous applications, \eg, movie trailer generation~\cite{irie2010autotrailer}
and narrative generation~\cite{grabska2021application}. Typical approaches of video summarization extract essential clips or frames from a given long video \cite{song2015tvsum,gygli2014creating,de2011vsumm}. Alternatively, the principal video content can also be summarized in natural language, e.g., video captioning~\cite{zhou2018towards,xu2016msr,krishna2017dense}. However, previous works treat either visual or textual summarization as separate tasks and thus ignore the semantic correlation between these two modalities of summarization. Therefore, these methods lack the ability to generate aligned visual textual summaries. An earlier attempt \cite{chen2017video} seeks to simultaneously generate visual and textual summaries from long videos. Still, the generated visual textual summaries in this work are not guaranteed to be semantically aligned since the two tasks were treated as separate, and there were no paired video and text summarization data for training or testing.

\begin{figure*}
    \centering
    \includegraphics[width=1.0\textwidth]{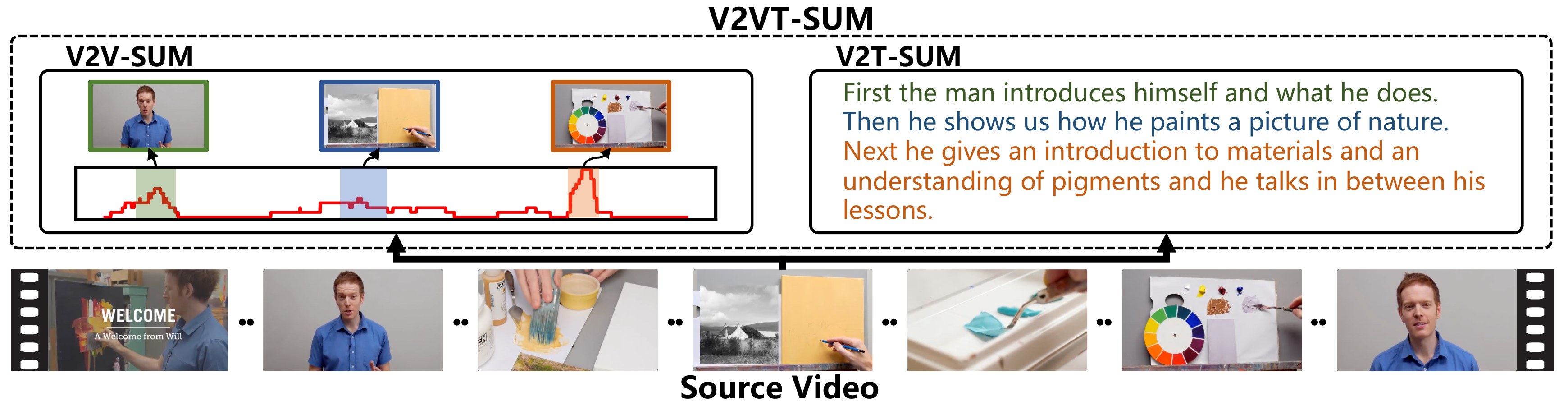}
    \vspace{-5mm}
    \caption{Illustration of the V2X-SUM tasks. A full-length source video ({\em bottom}) can be summarized into a shortened video and a text narrative ({\em top}). This task requires semantic alignment between the video and text summaries.}
    \vspace{-2mm}
    \label{fig:v2xsum}
\end{figure*}

In this study, we first introduce a novel cross-modal video summarization task, which involves generating visual and textual summaries with semantic coherence. To facilitate this new task, we propose \textbf{VideoXum}, an enriched large-scale dataset for cross-modal video summarization. The dataset is built on ActivityNet Captions \cite{krishna2017dense}, a large-scale public video captioning benchmark consisting of 200 distinct activity categories. These activity classes belong to 5 different top-level video topics: ``Eating and Drinking", ``Sports, Exercises, and Recreation", ``Socializing, Relaxing, and Leisure", ``Personal Care", and ``Household". To ensure consistent annotations, we hire workers to annotate ten shortened video summaries for each long source video according to the corresponding captions. Consequently, VideoXum contains 14K long videos with 140K pairs of aligned video and text summaries. Our goal is to extend the traditional single-modal video summarization task to a cross-modal video summarization task. Fig.~\ref{fig:v2xsum} presents this novel task termed V2X-SUM (Video-to-X Summarization), where X denotes the modality of generated summaries. According to the target modality, we categorize the V2X-SUM task into three subtasks:

\noindent \textbf{Video-to-Video Summarization (V2V-SUM)}. 
This task requires models to identify key segments from a source video and produce an abridged version.

\noindent \textbf{Video-to-Text Summarization (V2T-SUM)}.
In this task, models need to summarize the main content of the source video into a brief text description.

\noindent \textbf{Video-to-Video{\&}Text Summarization (V2VT-SUM)}. This task requires models to achieve V2V-SUM and V2T-SUM tasks simultaneously. Moreover, the semantics of these two modalities of summaries should be well aligned.

Compared with single-modal summarization tasks, cross-modal video summarization comes with its own challenges.
We summarize three primary challenges for this new task.
First, 
the scarcity of large-scale, diverse, and well-annotated cross-model video summarization benchmarks presents a significant hurdle for researchers in promoting the corresponding techniques.
Second, from the perspective of optimization, it is nontrivial to ensure the stability of the training process that accommodates both tasks concurrently. Specifically, a stable training process could facilitate learning of each single-modal task, thereby improving the overall performance.
Third, either assuring or evaluating the semantic coherence between the generated video and text summaries is challenging.

To establish strong baseline models for this emerging task, we propose VTSUM-BLIP, an end-to-end cross-modal video summarization model. 
To leverage the strong capability of vision-language pretrained (VLP) models for vision understanding and language modeling, we employ BLIP \cite{li2022blip} as our foundational backbone. This VLP encoder-decoder architecture provides a superior initialization, which is crucial for stable and effective optimization in machine learning models~\cite{zoph2020rethinking}.
Inspired by efficient video encoding techniques \cite{li2020hero,beltagy2020longformer,ju2022prompting,ni2022expanding}, we design an efficient hierarchical video encoding strategy, incorporating a frozen encoder, a temporal modeling module, and a context aggregation module to encode long videos. The video encoder is followed by different task-specific decoders for video and text summarization. The modularized design enables us to perform more complex downstream tasks without changing the structure of the pretrained backbone. 
Existing multimodal-based video summarization works~\cite{chen2017video,narasimhan2021clip} follow a pipeline where a summary is first generated in one modality, and then this generated summary is served as a prompt to improve the summary in another modality. Such methods may suffer from bias accumulation issues since they do not consider the semantic coherence of summaries in two modalities.
In contrast, our proposed VTSUM-BLIP enables joint training of the video and text summarization decoders in parallel. 
In other words, the predictions of these two decoders avoid sequential dependency between the two modalities of summaries.
Furthermore, the video and text summarization decoders collaboratively influence the shared parameters during training, allowing the framework to learn the semantic coherence between two tasks.

Our proposed framework achieves promising performance on VideoXum, as well as other existing single-modal video summarization datasets (\ie, TVSum~\cite{song2015tvsum}, SumMe~\cite{gygli2014creating}, and ActivityNet Captions~\cite{krishna2017dense}).
Inspired by the CLIPScore~\cite{hessel2021clipscore} and its video-text variant~\cite{wu2021godiva}, we adapt these metrics for the VideoXum and propose VT-CLIPScore for evaluating the semantic coherence of cross-modal summaries. The empirical results show the consistency of the proposed metric with human evaluation.

Our main contributions can be summarized as follows: 

\begin{itemize}
    \item We introduce \textbf{VideoXum}, an enriched large-scale dataset, to bridge the modality gap between the video and text summarization. The dataset contains 14K long videos with corresponding human-annotated video and text summaries. We conduct comprehensive experimental analyses to verify the rationality of our proposed new dataset. 
    \item Based on VLP encoder-decoder architecture, we propose an end-to-end cross-modal video summarization framework -- VTSUM-BLIP to establish strong baseline models for this novel task. The models achieve promising results on VideoXum and the new state of the art on several existing single-modal video summarization datasets.
    \item We propose an evaluation metric VT-CLIPScore on the VideoXum benchmark to evaluate cross-modal semantic consistency. The empirical results show the high consistency of our proposed metric with human evaluation.
\end{itemize}
\vspace{-2mm}

\section{Related Work}

\subsection{Video Summarization}
 Video summarization datasets (\eg,  SumMe~\cite{gygli2014creating}, TVSum~\cite{song2015tvsum}, and YouTube~\cite{de2011vsumm}) have enabled the development of state-of-the-art video summarization methods~\cite{narasimhan2021clip,zhang2016video,zhou2018deep,park2020sumgraph,saquil2021multiple}. Among these models, vsLSTM \cite{zhang2016video} first attempted to learn frame importance by modeling the temporal dependency among frames using LSTM \cite{graves2012long} units. The model can be combined with a determinantal point process (DPP) to improve the diversity of generated video summary. Following vsLSTM, several other approaches were proposed to model the temporal dependency, e.g., H-RNN~\cite{zhao2017hierarchical}, HSA-RNN\cite{zhao2018hsa}, DASP~\cite{ji2020deep}. Another solution models the spatiotemporal structure of the video to learn frame importance, such as MerryGoRoundNet~\cite{lal2019online}, and CRSum \cite{yuan2019spatiotemporal}. Adversarial learning-based methods~\cite{fu2019attentive,zhang2019dtr} can also perform well. Recently, multimodal-based video summarization method~\cite{narasimhan2021clip} leverages generated text summaries to promote predictions of frame-level scores for video summaries. Different from multimodal-based video summarization, the cross-modal video summarization task requires simultaneously producing both visual and textual summaries from a source video, which goes beyond generic video summarization. Moreover, it ensures semantic coherence between these two modalities.

\vspace{-2mm}
\subsection{Video Captioning}
Video Captioning aims to describe a video with text, which requires the capability of understanding actions and events.
Existing benchmarks (e.g., MSVD~\cite{chen2011collecting}, YouCook~\cite{zhou2018towards}, MSR-VTT~\cite{xu2016msr}, and ActivityNet Captions \cite{krishna2017dense}) have helped to promote the ability of language models to generate reasonable captions for video. Benefiting from these human-annotated datasets, many novel approaches are proposed. 
Attention-based methods~\cite{yao2015describing,yan2019stat} employ attention mechanisms to help the model in associating relevant frames since not every frame in a video is equally important.
DENSE~\cite{krishna2017dense} is an early attempt at dense video captioning, which detects events with an event proposal module and associates them with LSTM. Wang et al.~\cite{wang2018bidirectional} develop a bidirectional process to encode context for detecting event proposals. Moreover, Masked Transformer~\cite{zhou2018end} proposes a differentiable masking scheme to ensure consistency between event proposal and caption generation modules.

\vspace{-2mm}
\subsection{Multimodal Pretraining}
Large language models (LLMs) \cite{brown2020language,devlin2018bert,lewis2019bart,raffel2020exploring} have revolutionized NLP research in recent years. Following the large-scale pretraining models in the field of NLP, numerous works \cite{ju2022prompting,kim2021vilt,wang2020minilm,xue2021probing,zhang2021vinvl,hu2022promptcap} on exploring the combination of vision and language (VL) pretraining have achieved great success. Since then, image-text pretraining has become
a default approach to tackling VL tasks \cite{biten2019scene,lin2014microsoft,regneri2013grounding,singh2019towards}. In addition, the introduction of Vision Transformers \cite{dosovitskiy2020image} enables vision and language modalities to be jointly modeled by Transformers in a more scalable fashion \cite{alayrac2022flamingo,wang2022git,yu2022coca,yuan2021florence}. According to the encoding strategies for image and language modalities, VL models can be categorized into fusion encoder \cite{li2019visualbert,lu2019vilbert,su2019vl,tan2019lxmert}, dual encoder \cite{radford2021learning}, and a combination of both \cite{bao2021vlmo,du2022survey,singh2022flava}. Several video-language pretrained models have also shown strong performance on video captioning and other video tasks, such as HERO \cite{li2020hero}, VideoBERT \cite{sun2019videobert}, and UniVL \cite{luo2020univl}.
In this work, cross-modal video summarization requires models with strong video understanding and language modeling capabilities. Therefore, this new task provides a practical scenario to assess the superiority of multimodal pretrained models.

\begin{table*}[htbp]
\caption{Comparison with existing single-modal video-to-video summarization and video-to-text summarization datasets.}
\vspace{-2mm}
\centering
\begin{adjustbox}{width=0.98\textwidth}\setlength\tabcolsep{10pt}
\begin{tabular}{lllccccc}
\toprule[1pt]
\multirow{2}{*}{\bf Dataset} & \multirow{2}{*}{\bf Domain} & \multirow{2}{*}{\bf \# Videos} & {\bf  Avg Ratio (\%)} & {\bf Avg Len}  & \multicolumn{3}{c}{\bf Supported Task}                                                      \\
                         &                         &                            & {\bf VideoSum}      & {\bf TextSum} & \multicolumn{1}{c}{\bf V2V-Sum} & \multicolumn{1}{c}{\bf V2T-Sum} & \multicolumn{1}{c}{\bf V2VT-Sum} \\ \shline
MSVD~\cite{chen2011collecting}    &   open  &  1,970 & -              & 8.7     &                            {\scriptsize \XSolidBrush}     &          \Checkmark                   &                      {\scriptsize \XSolidBrush}                         \\
YouCook~\cite{zhou2018towards}    &   cooking  &  88 & -              & 15.9     &                            {\scriptsize \XSolidBrush}     &          \Checkmark                   &                      {\scriptsize \XSolidBrush}                        \\
MSR-VTT~\cite{xu2016msr}                   & open                 & 7,180                        & -              & 18.6       &                           {\scriptsize \XSolidBrush}     &          \Checkmark                   &                      {\scriptsize \XSolidBrush}                          \\
ActivityNet~\cite{krishna2017dense}               & open                    & 20,000                        & -              &  40   &                        {\scriptsize \XSolidBrush}     &          \Checkmark                   &                      {\scriptsize \XSolidBrush}       \\ \hline
SumMe~\cite{gygli2014creating}                     & 3 categories                      & 25                         &   $ 15\% <$            & -        &                         \Checkmark         &               {\scriptsize \XSolidBrush}              &    {\scriptsize \XSolidBrush}   \\
Youtube~\cite{de2011vsumm}                     & 5 categories                      & 50                         &   $ 15\% <$            & -        &                         \Checkmark         &               {\scriptsize \XSolidBrush}              &    {\scriptsize \XSolidBrush}   \\
TVSum~\cite{song2015tvsum}                     & 10 categories                      & 50                         &  $15\% <$             & -        &                    \Checkmark         &               {\scriptsize \XSolidBrush}              &    {\scriptsize \XSolidBrush}                         \\ \hline
{\bf VideoXum} (ours)     & open                    & 14,001                       & $13.6\%$         & 49.9     &                         \Checkmark    &                     \Checkmark        &   \Checkmark                          \\ \bottomrule[1pt]
\end{tabular}
\label{tab:datasets}
\end{adjustbox}
\vspace{-4mm}
\end{table*}

\section{Dataset}
\label{sec:dataset}

In this section, we introduce the proposed VideoXum dataset. The dataset is reannotated by a limited number of workers, including 14,001 long videos with 140,010 video and text summaries pairs. We describe the process of dataset collection and annotation strategy. We also provide several quantitative and qualitative analyses of the proposed dataset. Finally, we compare the VideoXum with existing single-modal video summarization datasets.

\subsection{Dataset Curation}

\begin{figure*}[t]
    \centering
    \captionsetup[subfloat]{labelfont=small,textfont=scriptsize}
    \vspace{-4mm}
    \subfloat[]{\includegraphics[width=0.24\textwidth]{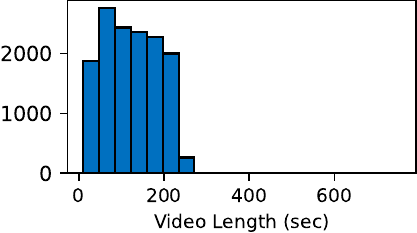}\label{fig:video_len}}
    \hfill
    \subfloat[]{\includegraphics[width=0.24\textwidth]{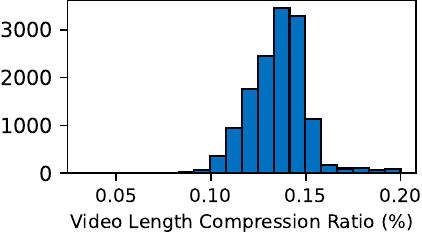}\label{fig:comp_ratio}}
    \hfill
    \subfloat[]{\includegraphics[width=0.235\textwidth]{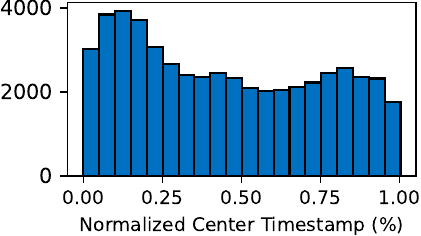}\label{fig:norm_center_hist}}
    \hfill
    \subfloat[]{\includegraphics[width=0.235\textwidth]{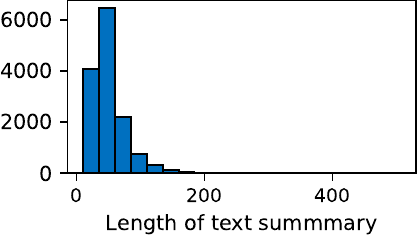}\label{fig:tsum_len}}
    \vspace{-2mm}
    \caption{Statistical information of VideoXum dataset: (a) distribution of video length; (b) distribution of video length compression ratio;  (c) distribution of normalized center timestamp;  (d) distribution of length of text summary.}
    \vspace{-4mm}
    \label{fig:dataset_hist}
\end{figure*}

\noindent \textbf{Dataset Collection}.
The VideoXum dataset is built based on ActivityNet Captions~\cite{krishna2017dense}, a high-quality public video captioning benchmark. There are three primary reasons to build upon ActivityNet Captions.
First, the dataset contains 20K real-life Youtube videos with diverse content, in terms of rich topics, different photographic devices, multiple view angles, and so on. Each video in this dataset is annotated with a series of dense temporal segments, and each segment corresponds to a concrete sentence description, offering diverse patterns essential for video understanding and generation tasks.
Second, the dataset contains numerous lengthy videos in Fig.~\ref{fig:video_len}, which introduces more challenges to the cross-modal video summarization task.
Third, as described in Section \ref{sec:intro}, the well-annotated sentence narratives are natural summaries of the source videos.
Therefore, the content and length of videos in the ActivityNet Captions dataset largely meet our requirements and provide an ideal foundation for constructing our cross-modal video summarization benchmark. To maintain our focus on long videos, we filter out videos shorter than 10 seconds. 

\noindent \textbf{Dataset Reannotation}.
For each video, we expect the total length of its video summary to be bound to 15\% of the source video, along with a semantically aligned text summary. ActivityNet Captions~\cite{krishna2017dense}  already contains video captions with temporal segments for long videos. Therefore, we concatenate the caption sentences as a text summary for the long source video. 
However, the annotated video spans, which cover an average of 94.6\% of the source videos, are too long to be regarded as a video summary by themselves since video summaries need to be much more concise. Therefore, we reannotate the video spans and obtain an abridged version of video segments for better aligning with the sentence captions.

Due to the inherently subjective nature of summarizing a long video (this conclusion is also reflected by the human performance on V2V-SUM in TABLE \ref{tab:videoxum_table}), it is hard to obtain perfect ground truth labels for this task.
Following previous works~\cite{song2015tvsum,lei2021detecting,sun2014ranking}, we required ten different workers to annotate video summary spans corresponding to a same text description. For each given caption, we obtained ten shortened spans. During the evaluation, we compared the prediction with all ten annotations and then obtain the average score for each video.
To further ensure consistent annotations, we hired 40 workers in total to reannotate all 140,010 summarised video spans over a period of two months. On average, each worker reannotated about 15 videos per hour. To maintain high-quality annotations, we regularly reviewed the reannotated video spans and provided feedback to workers. Every 24 hours, we randomly evaluated 15\% of an annotation batch for accuracy. If the acceptance rate of the sampled annotations reached 90\%, we considered the entire annotation batch as passed; otherwise, we asked workers to reannotate the batch.

This reannotation pipeline aims to obtain an abridged version (ideally bounded to 15\%) of videos for better aligning with the sentence captions. Therefore, we filter the initial ActivityNet dataset using the length compression ratio of video with 20\% as the threshold. The video length compression ratio is calculated as $\text{Ratio}(S, V) = \frac{|S|}
{|C|}$, where $|S|$ denotes the length of summary, $|C|$ denotes the length of source video. Finally, 14,001 long videos remain in our dataset.

\noindent \textbf{Dataset Split}.
We split the dataset into training, validation, and test sets. The split strategy also guarantees that all three data splits preserve the same distribution of video length. In particular, the dataset is divided into 8,000, 2,001, and 4,000 videos in the training, validation, and test sets, respectively.

\vspace{-2mm}
\subsection{Dataset Statistics}
Fig.~\ref{fig:dataset_hist} presents the statistical information of the VideoXum dataset. As shown in Fig.~\ref{fig:video_len}, it shows that the length of the videos ranges from 10 to 755 seconds, with 99.9\% of them under 300 seconds. The average length is 124.2 seconds, and the median length is 121.6 seconds.
For the video summarization task, most video summary lengths are shorter than 15\% of the source video length. Fig.~\ref{fig:comp_ratio} shows that the average length compression ratio is 13.6\%, with a median ratio of 13.7\%, and a maximum ratio of 20\%.
Moreover, we investigate the distribution of the center timestamps of important clips. All the center timestamps are normalized to fall within the range of $[0, 1.0]$ according to the original video length. Fig.~\ref{fig:norm_center_hist} suggests that the important clips are generally uniformly distributed throughout the video, with a mild peak at the beginning. Therefore, the VideoXum dataset does not suffer from temporal bias issues~\cite{lei2021detecting}.
For the text summarization task, each video is summarized into a narrative paragraph that describes multiple events. On average, each narrative paragraph contains 49.9 words. Fig.~\ref{fig:tsum_len} indicates that most (98\%) text summaries are shorter than 128 words, which guides us to set the maximum text generation length as 128.

\begin{table}[]
\small
\centering
\caption{Comparison of F1 score on {\it human annotations}. F1-avg denotes the averaged F1 score across all reference summaries. F1-max represents the maximum F1 score. Symbol $^\sharp$ denotes the results directly quoted from \cite{otani2019rethinking}.}
\vspace{-1mm}
\begin{adjustbox}{width=1\linewidth}\setlength\tabcolsep{7pt}
\begin{tabular}{cccccc}
\shline
\multicolumn{2}{c}{\textbf{VideoXum (ours)}} & \multicolumn{2}{c}{\textbf{SumMe}} & \multicolumn{2}{c}{\textbf{TVSum}} \\
\textbf{F1-Avg}  & \textbf{F1-Max} & \textbf{F1-Avg}  & \textbf{F1-Max} & \textbf{F1-Avg}         & \textbf{F1-Max}         \\ \hline
36.2                    & 59.5 & 31$^\sharp$               & 54$^\sharp$              & 54$^\sharp$               & 78$^\sharp$                               \\ \shline
\end{tabular}
\end{adjustbox}
\label{tab:human_eval}
\vspace{-3mm}
\end{table}

\subsection{Comparison with Existing Single-modal Video Summarization Datasets}
In TABLE~\ref{tab:datasets}, we compare the proposed VideoXum dataset with existing {\it single-modal} video-to-video and video-to-text\footnote{In this paper, we regard the video captioning task as video-to-text summarization task} summarization datasets.
The main difference between VideoXum and other existing datasets is that VideoXum contains aligned human-annotated video and text summaries, while others only have single-modal summaries for source videos.
Compared with the existing video summarization benchmarks (e.g., SumMe~\cite{gygli2014creating} and TVSum~\cite{song2015tvsum}), the amount of data in the VideoXum dataset is significantly larger. In addition, VideoXum contains open-domain videos with more diverse scenarios than other datasets. To ensure the quality of human annotation, we evaluate annotated data using a leave-one-out strategy~\cite{otani2019rethinking}. TABLE~\ref{tab:human_eval} shows that our annotation quality is comparable with existing benchmarks.

\section{Methodology}

\begin{figure*}
    \centering
    \includegraphics[width=1\textwidth]{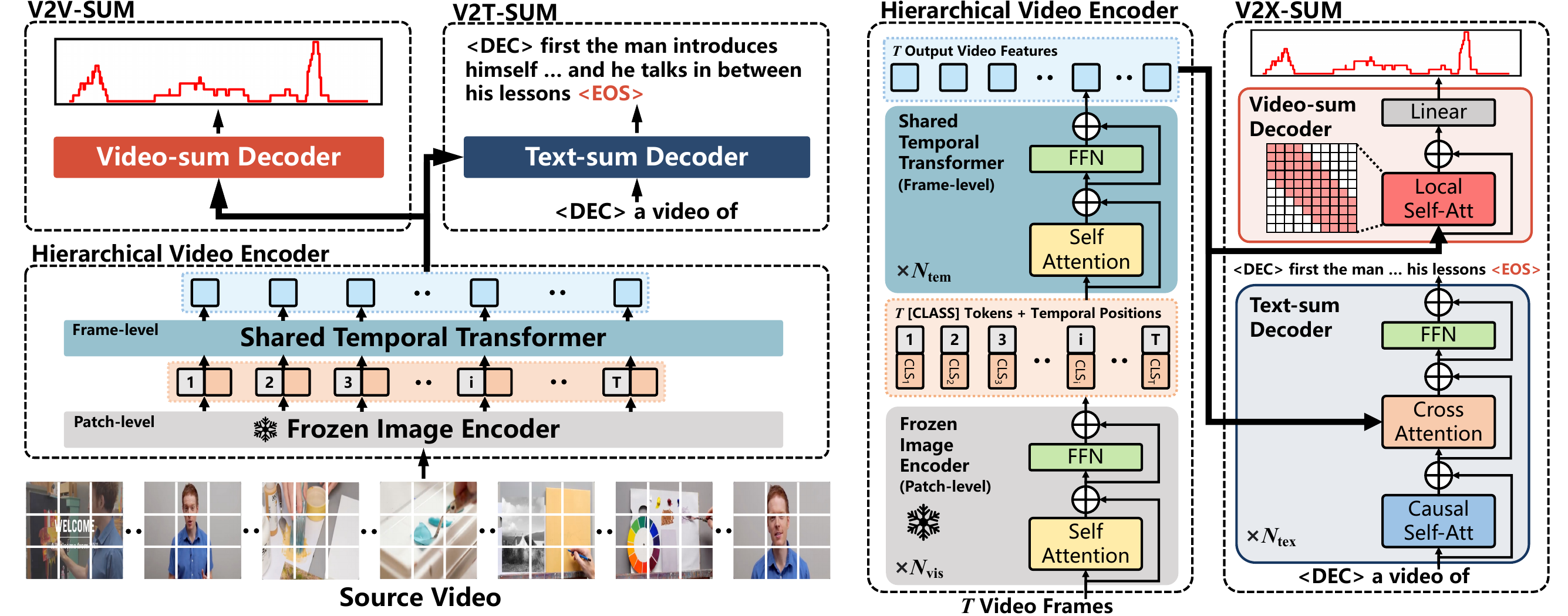}
    \vspace{-6mm}
    \caption{An overview of our VTSUM-BLIP framework ({\em left}). It consists of a hierarchical video encoder ({\em middle}), video-sum decoder, and text-sum decoder ({\em right}). For V2V-SUM, the video-sum decoder employs a temporal Transformer and local self-attention module to aggregate the local context. For V2T-SUM, the text-sum decoder is a pretrained BLIP text decoder.}
    \label{fig:v2xsum-framework}
    \vspace{-4mm}
\end{figure*}

\subsection{Problem Formulation}
 We formulate the problem of cross-modal video summarization as a multi-task learning problem, including V2V-SUM and V2T-SUM. Given a video $\mathcal{V}=\{\bv_{i}\}_{i=1}^{T}$,  $T$ is the number of frames in the video and $\bv_i$ denotes the $i\text{-th}$ frame in the temporal order.
 Our goal is to learn a shared video encoder $f(\cdot; \theta)$ followed by two task-specific decoders, including a video summarization (video-sum) decoder $g_v(\cdot; \theta_{v})$ and a text summarization (text-sum) decoder $g_t(\cdot; \theta_t)$. In particular, the notations of $\theta$, $\theta_v$, and $\theta_t$ represent the learnable parameters of the shared video encoder, video-sum decoder, and text-sum decoder, respectively. We first feed the input video $\mathcal{V}$ into the shared video encoder to produce the video features $\tilde{\mathcal{Z}}$:
\begin{align}
    \tilde{\mathcal{Z}} &= f(\mathcal{V}, \mathcal{E}_\text{temp}; \theta), \label{eq:vis_feat}
\end{align}
where $\mathcal{E}_\text{temp}$ is the temporal position embedding for the video frames. Given the video features $\tilde{\mathcal{Z}}$, the model generates a video summary $\mathcal{V}_\text{sum}$ and a text summary $\mathcal{T}_\text{sum}$. In particular, we can formulate visual and narrative outcomes as follows:
\begin{align}
    \mathcal{V}_\text{sum} &= g_v(\tilde{\mathcal{Z}}; \theta_v),  \label{eq:vsum} \\
    \mathcal{T}_\text{sum} &= g_t(\tilde{\mathcal{Z}}, \mathcal{T}_\text{prompt}; \theta_t),  \label{eq:tsum}
\end{align}
where $\mathcal{T}_\text{prompt}$ denotes a prompt sequence.

\vspace{-2mm}
\subsection{Cross-modal Video Summarization}
Our proposed VideoXum benchmark requires models with strong video understanding and language modeling capabilities. To this end, we employ the large vision-language pretrained (VLP) model BLIP~\cite{li2022blip} as our backbone.
For efficient video encoding, we propose a hierarchical video encoder to capture spatiotemporal features. Followed by the video encoder, video-sum and text-sum decoders are designed for V2V-SUM and V2T-SUM tasks, respectively.
The overall framework termed VTSUM-BLIP is shown in Fig.~\ref{fig:v2xsum-framework}.

\noindent \textbf{Hierarchical Video Encoder}. 
The hierarchical video encoder $f(\cdot; \theta)$ aims to address the challenge of efficiently extracting spatiotemporal visual features from a long video.
Drawing the inspiration from efficient video encoding~\cite{li2020hero,beltagy2020longformer,ju2022prompting,ni2022expanding} and long document summarization \cite{ruan2022histruct+}, we formulate the BLIP image encoder into a hierarchical architecture for long video encoding without changing the structure of the encoder. This enables us to efficiently obtain rich video features at both video frame and image patch levels.
Specifically, given a video $\mathcal{V}=\{\bv_{i}\}_{i=1}^{T}$ with $T$-frame, the frozen image encoder projects each video frame $\bv_i$ into the representation space and produce $T$ visual tokens $\mathcal{Z}=\{\bz_{i}\}_{i=1}^{T}$. Next, we use temporal position embedding  $\mathcal{E}_\text{temp}=\{\bolde_i \}^T_{i=1}$ with the shared \textbf{T}emporal \textbf{T}ransformer (TT) to model the temporal information for the video sequence. In this way, we can obtain spatiotemporal visual features $\tilde{\mathcal{Z}}=\{\tilde{\bz_{i}}\}_{i=1}^{T}$.

To better understand the hierarchical video encoder, we break it down into two key components:
\begin{itemize}
    \item \textbf{Frozen Image Encoder}.
    Following the previous works~\cite{ju2022prompting,zhang2021vinvl,chen2020uniter}, we freeze the parameters of the pretrained BLIP encoder, which can help to improve the training time and GPU memory efficiency for encoding long videos. In detail, we first convert input images into several patches as the input tokens for the $N_\text{vis}$-layer BLIP encoder. The patch embedding is prepended with a [\texttt{CLS}] token in the representation space. Next, we take all output of the [\texttt{CLS}] tokens as the representation of the input frames. We can compress the input video at the frame level through the hierarchical encoding strategy and generate the representation $\mathcal{Z}$.

    \item \textbf{Shared Temporal Transformer}.
    After obtaining a sequence of the video frame representation $\mathcal{Z}=\{\bz_i\}^T_{i=1}$, we add these temporal position embeddings $\mathcal{E}_\text{temp}=\{\bolde_i \}^T_{i=1}$ to $\mathcal{Z}$, and feed them into the shared temporal Transformer (TT) for temporal modeling and get the spatiotemporal visual features $\tilde{\mathcal{Z}}=\{\tilde{\bz _i}\}^T_{i=1}$ in Eq.(\ref{eq:vis_feat}): \vspace{-4mm}
    \begin{align}
        \nonumber &\bz_i^{(0)} = \bz_i+\bolde_i, \\ 
        \nonumber &\bz_i^{(l)} = \text{TT}^{(l)}(\bz_1^{(l-1)},\dots,\bz_T^{(l-1)}), \ l=1,\dots,N_{tem},\\
        &\tilde{\bz_i} = \bz_i^{(N_{tem})},
    \end{align}
    where $l$ indicates the $l$-th block of the temporal Transformer, and $N_{tem}$ denotes total block number of the temporal Transformer.
\end{itemize}

\noindent \textbf{Video-Sum Decoder}.
Inspired by long document encoding technique~\cite{beltagy2020longformer}, the video-sum decoder $g_v(\cdot; \theta_{v})$ employs a \textbf{C}ontext \textbf{A}ggregation (CA) module that captures context from neighboring frames with local self-attention.
In particular, we first define a fixed-size slice window at each temporal position and then construct a binary local attention map ${M}^{LA} \in \{0,1\}^{T \times T}$ with a given window size $\varepsilon$. For example,  Fig.~\ref{fig:v2xsum-framework} ({\em right}) presents a local attention map with a window size $\varepsilon=7$. Next, we compute the local attention features $\mathcal{A}_{loc}$:\vspace{-1mm}
\begin{align}
    \mathcal{A}_{loc} = \left(\text{softmax}(\frac{QK^T}{\sqrt{d}}) \odot M^{LA}\right)V,
\end{align}
where $\odot$ is element-wise multiplication, and queries $Q$, keys $K$, and values $V$ are $d$-dimensional features generated from temporal-aware visual features $\tilde{\mathcal{Z}}$. Finally, we feed local attention-enhanced features into a linear classifier to obtain the predictions of the frame-level importance scores $\{p_i\}_{i=1}^T$. 
Our training objective for video summarization is an averaged binary cross-entropy loss, as following:
\begin{equation}
\small
    \mathcal{L}_{v} = -\frac{1}{T}\sum^{T}_{i=1} (\hat{y}_i\cdot \log(p_i) + (1 - \hat{y}_i)\cdot \log(1- p_i)),
\end{equation}
where $\hat{y_i} \in \{0,1\}$ denotes whether the $i$-th frame is a key frame, and $p_i$ indicates the predicted importance score of the $i$-th frame. Finally, we select the top 15\% of frames to attain a video-sum result $\mathcal{V}_\text{sum}$ in Eq.(\ref{eq:vsum}) from a long video. 

\noindent \textbf{Text-Sum Decoder}.
The pretrained BLIP text decoder is a strong baseline for
text generation. The text summarization decoder $g_t(\cdot; \theta_{t})$ contains $N_{tex}$ stacked Transformer decoder blocks with cross-attention modules. During the decoding process, the text decoder takes a prompt sequence $\mathcal{T}_\text{prompt}$, 
 and the video features $\tilde{\mathcal{Z}}=\{\tilde{\bz_i}\}^T_{i=1}$ from the video encoder as inputs and then generate the final text summary $\mathcal{T}_\text{sum}$ in Eq.(\ref{eq:tsum}).
The training objective of text summarization is negative log-likelihood (NLL), which can be expressed in the equation as:
\begin{equation}
    \mathcal{L}_{t} = - \sum^{N_{tex}}_{i=1} \log P(w_i\vert w_0, w_1,\dots, w_{i-1}, \tilde{\mathcal{Z}}),
\end{equation}
where $w_i$ denotes the $i$-th word in the sentence, $N_{tex}$ is the length of output sequence.

\vspace{-3mm}
\subsection{Overall Objective}
Following the multi-task learning paradigm, the overall objective of our proposed framework is calculated as the integration of video-sum loss $\mathcal{L}_v$ and text-sum loss $\mathcal{L}_t$:
\begin{equation}
    \mathcal{L} = \lambda_v \mathcal{L}_v + \lambda_t\mathcal{L}_t,\label{eq:loss}
\end{equation}
where $\lambda_v$ and $\lambda_t$ are the weights of different summary tasks.

\section{Experiments}
In this section, we first introduce the baseline models and experimental setup for the proposed VideoXum dataset. Then, we present the evaluation metrics and human evaluation strategy. In addition, we report several baseline models' performances under different settings and present a comprehensive experimental analysis to prove the effectiveness of our proposed method.

\subsection{Baseline Models}

We introduce all the baseline models listed in TABLE~\ref{tab:videoxum_table}:

\noindent {\bf Frozen-BLIP} refers to inference over the test set using a frozen BLIP model without training. We take this zero-shot setting performance as a lower bound for our benchmark. 

\noindent {\bf VSUM-BLIP (Base)} is a baseline model to perform video-to-video summary. It consists of a frozen BLIP encoder and a learnable video-sum decoder.

\noindent {\bf TSUM-BLIP (Base)} is a video-to-text summary baseline. We employ the vanilla BLIP model with a frozen encoder.

\noindent {\bf VTSUM-BLIP (Base)} combines the VSUM-BLIP and TSUM-BLIP modules of the model. It is comprised of a shared frozen encoder and two task-specific decoders. 

\noindent {\bf Temporal Transformer (TT)} is a crucial module to achieve the hierarchical encoding for videos while incorporating temporal information into a video sequence. Specifically, we use several Transformer layers combined with temporal positional embedding to model the temporal information.

\noindent {\bf Context Aggregation (CA)} is a plug-and-play module to model the video frame representations for the V2VSum task. Compared with the baseline models, this mechanism enhances the local context information for video representations and could help reduce the redundancy of video summaries.

\vspace{-3mm}
\subsection{Experimental Setup and Implementation Details} \label{sec:exp_setup}

\noindent \textbf{Data Preprocessing}. Video frames of all \textit{train/val/test} sets are first resized using bi-linear resampling to 224 pixels along the shorter side. Next, a $224 \times 224$ center crop is applied to the resized frames. This is a common preprocessing method. For each training batch, we add padding to all video sequences to make them the same length, enabling the videos to be processed in parallel and speeding up the training process. In addition, the padding tokens are masked out during the self-attention calculation. Based on data statistics in Fig.~\ref{fig:video_len}, we set the maximum video length to 512, and frames exceeding the maximum length are truncated.
For each text summary, we concatenate (dense) sentence captions in a video to construct a narrative paragraph~\cite{gabeur2020multi,patrick2021supportset,zhang2018cross}. According to data statistics in Fig.~\ref{fig:tsum_len}, we set the maximum generation length to 128 in the text summarization task.

\begin{table*}[t]
\centering
\caption{The performance of the baseline models on the VideoXum dataset {\em test} set for three different V2XSum tasks. The F1 score, BLEU@4, METEOR, ROUGE-L, CIDEr, and VT-CLIPScore are shown in \%.}
\vspace{-2mm}
\small
\begin{adjustbox}{width=0.99\textwidth}\setlength\tabcolsep{10pt}
\begin{tabular}{lcccccccc}
\toprule[1.2pt]
\multirow{2}{*}{Method} & \multicolumn{3}{c}{\textbf{V2V-SUM}} & \multicolumn{4}{c}{\textbf{V2T-SUM}} & \textbf{V2VT-SUM}   \\ \cmidrule(r){2-4} \cmidrule(r){5-8} \cmidrule(r){9-9}
                        & F1 score     & Kendall     & Spearman    & BLEU@4  & METEOR  & ROUGE-L  & CIDEr & VT-CLIPScore   \\ \midrule[1.2pt]
Frozen-BLIP                  & 16.1     & 0.008        & 0.011        & 0.0     & 0.4     & 1.4      & 0.0   & 19.5               \\  \midrule[1.2pt]
\multicolumn{9}{>{\columncolor[gray]{.85}}l}{\textbf{Single-Modal Video Summarization}}  \vspace{0.2em}                                                                             \\
\multicolumn{9}{>{\columncolor{myred}}l}{{Video-to-Video Summarization}}  \vspace{0.2em} \\
VSUM-BLIP (Base)          & 21.7 &	0.131 &	0.207        & -       & -       & -        & -     & -                 \\
$\ \ $+ Temporal Transformer  & 22.1 &	0.168 &	0.222          & -       & -       & -        & -     & -                 \\
$\ \ $+ Context Aggregation        & 22.2 &	0.172 &	0.228          & -       & -       & -        & -     & -                 \\
$\ \ $+ TT + CA           & 23.1 &	0.185 &	0.246        & -       & -       & -        & -     & -                 \\
\multicolumn{9}{>{\columncolor{myblue}}l}{{Video-to-Text Summarization}}  \vspace{0.2em} \\
TSUM-BLIP (Base)          & -        & -           & -           & 5.5 &	11.7 &	24.9 & 	18.6  & -                 \\ 
$\ \ $+ Temporal Transformer  & -        & -           & -          & 5.6      & 11.8      & 24.9      & 20.9    & -                \\ \midrule[1.2pt]
\multicolumn{9}{>{\columncolor[gray]{.85}}l}{\textbf{Cross-Modal Video Summarization}}    \vspace{0.2em} \\
{Two-stage Manner}          & {19.4} &	{0.107} &	{0.143}          &{5.1} &	{11.2} &	{24.0} & 	{15.6} & {28.2}                \\
VTSUM-BLIP (Base)          & 21.7 &	0.131 &	0.207          &5.5 &	11.7 &	24.9 & 	18.6 & 28.4                \\
$\ \ $+ Temporal Transformer  & 22.4       & 0.176          & 0.233          & 5.7      & 12.0      & 24.9       & 22.4    & 28.9                \\
$\ \ $+ Context Aggregation        & 22.2 &	0.172 &	0.228          & 5.5 &	11.7 &	24.9 & 	18.6     & 28.6                \\
$\ \ $+ TT + CA           & {\bf 23.5}    & {\bf 0.196}       & {\bf 0.258}      & {\bf 5.8}     & {\bf 12.2}    & {\bf 25.1}     & {\bf 23.1}  & {\bf 29.4}                \\ \midrule[1.2pt]
Human                   & 33.8 &	0.305 &	0.336 &       5.2       & 14.7       & 25.7        & 24.2     & 38.0                 \\
\bottomrule[1.2pt]
\end{tabular}
\end{adjustbox}
\vspace{-3mm}
\label{tab:videoxum_table}
\end{table*}

\noindent \textbf{Model Architecture}. We employ ViT-B/16~\cite{dosovitskiy2020image} as the image encoder backbone with $N_\text{vis}=12$ layers.
The $N_\text{tem}$-layer Temporal Transformer (TT) follows the image encoder, where $N_\text{tem}$ is 1. The temporal positional embeddings $\varepsilon_\text{temp}$ in Eq.(\ref{eq:vis_feat}) are also learnable.
The video-sum decoder contains a Context Aggregation (CA) module capturing local context and a binary linear classifier. The CA module constructs a binary local attention map with window size $\epsilon=5$.
For the text-sum decoder, we adopt a variant of Transformer with $N_\text{tex}=12$ layers, which replaces the bidirectional self-attention module with a causal self-attention module~\cite{dosovitskiy2020image}. In addition, the prompt $\mathcal{T}_\text{prompt}$ of the text-sum decoder in Eq.(\ref{eq:tsum}) is set as ``[\texttt{DEC}] a video of''.

\begin{table}[t]
\centering
\caption{Comparison with state-of-the-art methods on the TVSum and SumMe datasets.}
\vspace{-2mm}
\begin{adjustbox}{width=0.985\linewidth}\setlength\tabcolsep{8.5pt}
\begin{tabular}{lcccc}
\toprule[1.2pt]
\multicolumn{1}{c}{\multirow{2}{*}{Method}} & \multicolumn{2}{c}{\bf TVSum} & \multicolumn{2}{c}{\bf SumMe} \\
 \cmidrule(r){2-3} \cmidrule(r){4-5}
\multicolumn{1}{c}{}                        &  Kendall  & Spearman  & Kendall  & Spearman \\ \midrule[1.2pt]
dppLSTM~\cite{zhang2016video}                                     &     0.042     &    0.055      &     -     &       -   \\
DR-DSN~\cite{zhou2018deep}                                  &     0.020     &    0.026      &     -     &       -   \\
Sumgraph~\cite{park2020sumgraph}                                     &     0.094     &    0.138      &     -     &       -   \\
CLIP-it~\cite{narasimhan2021clip}                                     &     0.108     &    0.147      &     -     &       -   \\
Standard ranker~\cite{saquil2021multiple}                                         & 0.176    &   0.230        &  0.011   &   0.014         \\
\midrule[0.5pt]
VSUM-BLIP (Base)                                               &    0.160       &  0.207   &         0.154 & 0.191         \\
$\ \ $+ Temporal Transformer                                   &  0.182 &  0.239   &       0.266   &           0.330     \\
$\ \ $+ Context Aggregation                                   &  0.185 &  0.243   &       0.268   &           0.332     \\
$\ \ $+ TT + CA                                  &  {\bf 0.200} &  {\bf 0.261}   &       {\bf 0.295}   &           {\bf 0.365}     \\\bottomrule[1.2pt]
\end{tabular}
\end{adjustbox}
\label{tab:v2vsum_table}
\vspace{-3mm}
\end{table}

\noindent \textbf{Weight Initialization}. To initialize the weights of our model, we employ a state-of-the-art VLP model called BLIP~\cite{li2022blip}. The image encoder and the text-sum decoder are initialized by pretrained BLIP$_\text{CapFilt-L}$. Additionally, the Temporal Transformer and video-sum decoder are randomly initialized.

\noindent \textbf{Optimization}. Due to limited computational resources, we finetuned all of the parameters in our proposed VTSUM-BLIP model, except for the image encoder. We adopt the AdamW~\cite{loshchilov2018decoupled} optimizer with an initial learning rate of $2\times10^{-5}$ to optimize the model, and the $\beta_1=0.9, \beta_2=0.999$. The batch size is 64, and weight decay is $5\times10^{-2}$. The learning rate follows a cosine decay schedule~\cite{loshchilov2016sgdr} with the minimum learning rate of $0.0$. We train the VTSUM-BLIP framework for 56 epochs with a batch size of 64 on 4 A100 GPUs. In addition, the weights of video-sum loss $\mathcal{L}_v$ and text-sum loss $\mathcal{L}_t$ are $\lambda_v=15.0$ and $\lambda_t=1.0$, respectively.

\subsection{Evaluation}

\noindent \textbf{Video Summary Evaluation}. Following previous works~\cite{narasimhan2021clip,otani2019rethinking,saquil2021multiple} for video summarization evaluation, we adopt the F1 score, Kendall's $\tau$~\cite{kendall1945treatment}, and Spearman's $\rho$~\cite{zwillinger1999crc} as our automatic evaluation metrics.

\noindent \textbf{Text Summary Evaluation}. To evaluate the quality of generated text summaries for video {\bf text summary}, we adopt several metrics for video captioning evaluation \cite{xu2023mplug} including: BLEU~\cite{papineni2002bleu}, METEOR~\cite{banerjee2005meteor}, ROUGE-L~\cite{lin2004rouge}, CIDEr~\cite{vedantam2015cider}.

\noindent \textbf{Video-text Semantic Consistency Evaluation}. Apart from independently evaluating single-modal summaries, we also evaluate \textit{semantic consistency} of text and video summaries.
Inspired by previous works~\cite{hessel2021clipscore,wu2021godiva,singer2022make}, we adapt CLIPScore for VideoXum benchmark and introduce a new evaluation metric -- VT-CLIPScore for evaluating the text and video semantic consistency. Specifically, we finetune the vanilla CLIP model~\cite{radford2021learning} on VideoXum dataset with contrastive learning strategies.
It is worth noting that adapting CLIPScore to our proposed benchmark is necessary since there is a domain gap between the CLIP pretraining data and our VideoXum data. Therefore, finetuning the CLIP model on our data makes the evaluation score more reliable. Similar attempts of finetuning evaluation models (\ie, BERTScore~\cite{Zhang2020BERTScore} and Sentence-BERT~\cite{reimers2019sentence}) also support the necessity of the VT-CLIPScore.

To facilitate reimplementation, we use the AdamW optimizer with an initial learning rate of $2\times10^{-6}$ and a weight decay of $5\times10^{-2}$. We finetune the CLIP model for 50 epochs with a batch size of 16 on 4 GPUs. The empirical results in TABLE~\ref{tab:clipscore} show that our proposed VT-CLIPScore is sensitive enough to the semantic change of video and text. Moreover, the results in TABLE~\ref{tab:he} indicate the high consistency of our proposed automatic evaluation metric with human evaluation.

\begin{table}[t]
\small
\centering
\caption{Comparison with state-of-the-art methods on the ActivityNet Captions dataset.}
\vspace{-2mm}
\begin{adjustbox}{width=0.99\linewidth}\setlength\tabcolsep{5.3pt}
\begin{tabular}{lcccc}
\toprule[1.2pt]
\multicolumn{1}{c}{\multirow{2}{*}{Method}} & \multicolumn{4}{c}{\bf ActivityNet Captions} \\ \cmidrule(r){2-5}
 & BLEU@4 & METEOR & ROUGE-L & CIDEr \\ \midrule[1.2pt]
DENSE~\cite{krishna2017dense}     &    1.6    &    8.9    &     -    &    -   \\
DVC-D-A~\cite{li2018jointly}     &    1.7    &   9.3     &    -     &  -     \\
Wang \textit{et al.}~\cite{wang2018bidirectional}  &    2.3    &   9.6     &   
 19.1     &  12.7     \\
Bi-LSTM+TempoAttn~\cite{zhou2018end}     &      2.1  &    10.0    &  -       &   -    \\
Masked Transformer~\cite{zhou2018end}     &   2.8     &    11.1    &      -   &   -    \\
Support-Set~\cite{patrick2021supportset}     &    1.5    &    6.9    &      17.8   &    3.2   \\ \midrule[0.5pt]
TSUM-BLIP (Base) &  5.5  & {12.1} & 25.1 & 19.7    \\
$\ \ $+ Temporal Transformer                                      & {\bf 5.7}    &    {\bf 12.1}    &     {\bf 25.2}    &  {\bf 22.2}      \\\bottomrule[1.2pt]
\end{tabular}
\end{adjustbox}
\vspace{-4mm}
\label{tab:v2tsum_table}
\end{table}

\vspace{-2mm}
\subsection{Results on VideoXum}
\begin{figure}
    \centering
    \includegraphics[width=0.94\linewidth]{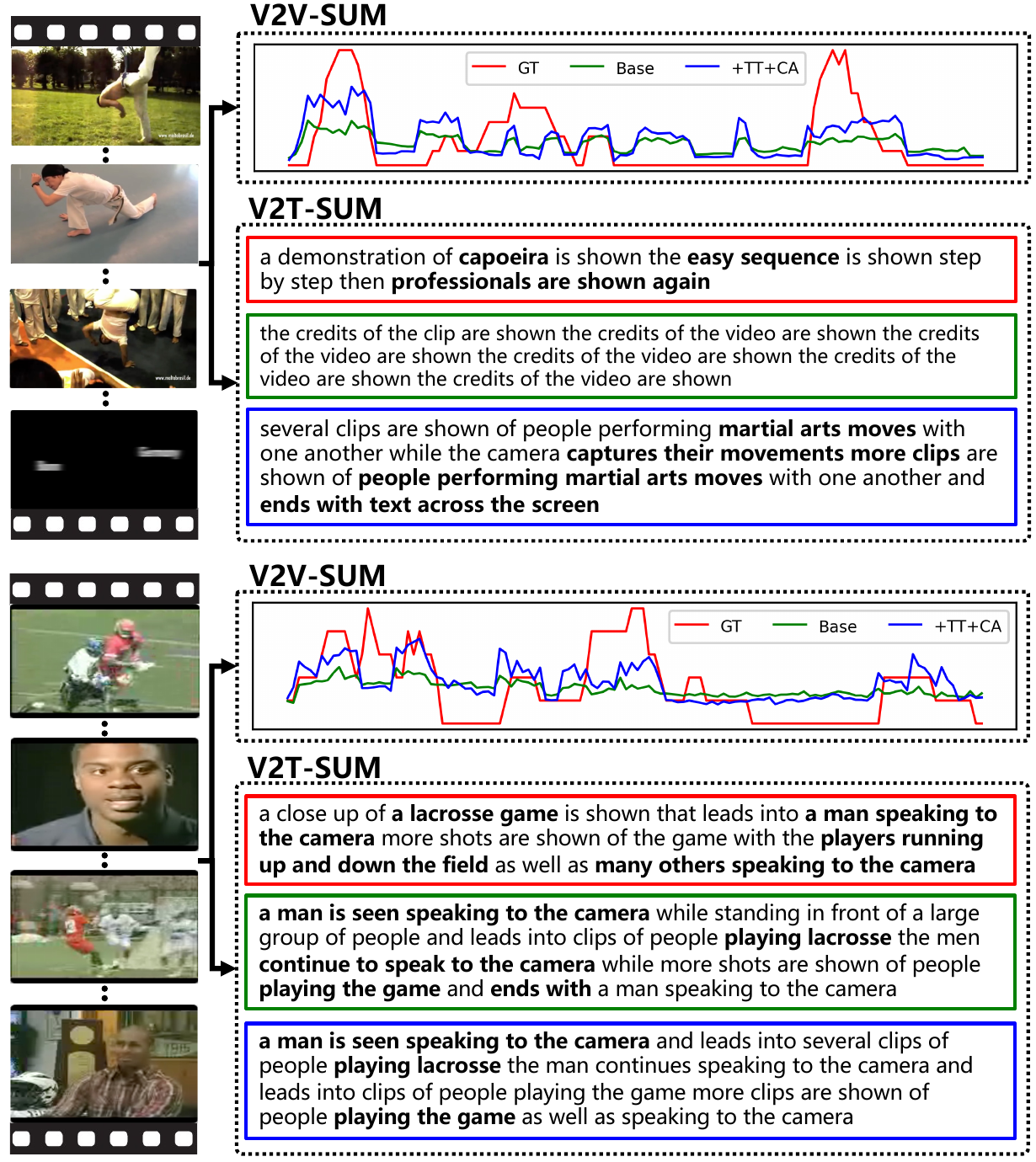}
    \vspace{-1mm}
    \caption{{\underline{Two}} example results of the generated video and text summaries across different baseline models. Red (both line and box) indicates the results of the ground truth. Green indicates the results of the VTSUM-BLIP (Base). Blue indicates the results of VTSUM-BLIP (+TT+CA).}
    \label{fig:vis}
    \vspace{-5mm}
\end{figure}

\begin{figure*}[t]
    \centering
    \includegraphics[width=0.98\textwidth]{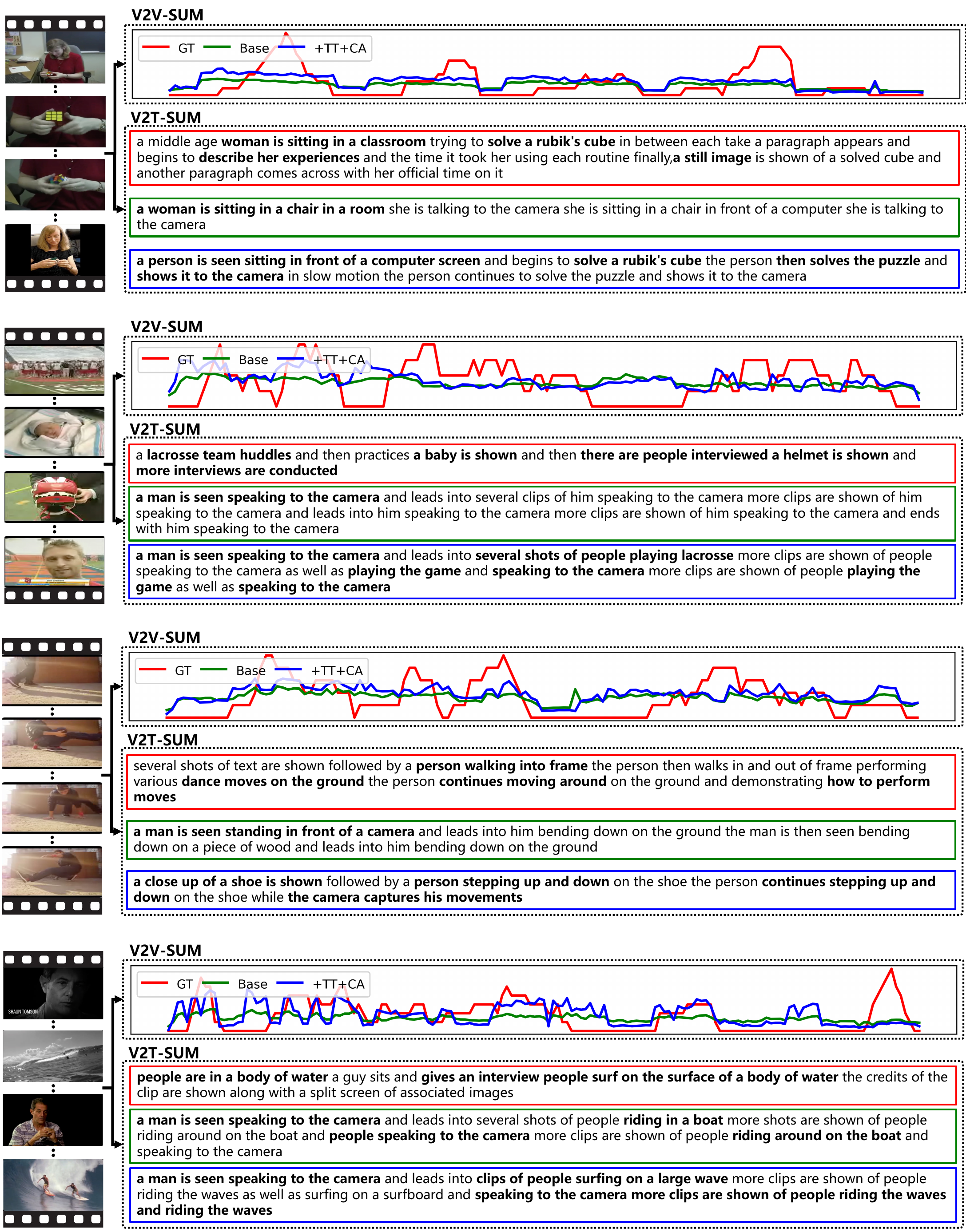}
    \caption{Additional example results of the generated video and text summaries across different baseline models. Red (both line and box) indicates the results of the ground truth. Green indicates the results of the VTSUM-BLIP (Base). Blue indicates the results of VTSUM-BLIP (+TT+CA).}
    \label{fig:add_vis}
\end{figure*}

We conduct experiments on VideoXum using different baseline models. TABLE~\ref{tab:videoxum_table} shows the empirical results of the models on VideoXum. By comparing VSUM-BLIP (Base), TSUM-BLIP (Base), and VTSUM-BLIP (Base) with Frozen-BLIP, BLIP models show better results after finetuning on specific tasks.
The comparison between the end-to-end VTSUM-BLIP (Base) and the Two-stage Manner~\cite{chen2017video} (\ie, first V2V-Sum and then V2T-Sum) demonstrates the superiority of the end-to-end framework since errors originating in the V2V-SUM stage could negatively influence the V2T-SUM stage.
In all three tasks, the model with TT can help model the video sequence better, indicating that temporal information is necessary. The CA module can enhance the local information awareness of the model, which can help improve the performance of the V2V-SUM task. The performance gains of TT on V2V-SUM are more significant than that on V2T-SUM, one of the possible reasons is that the text decoder is a well-generalized model trained on a large corpus and is insensitive to the subtle changes of the input features \cite{arora2018stronger,hua2021noise,hua2022fine}. A Base BLIP model combined with TT and CA achieves the SOTA results in our proposed three tasks. From the overall performance,  our proposed multitask framework can benefit both V2V and V2T-SUM tasks. In addition, TABLE~\ref{tab:videoxum_table} shows the human performance on VideoXum. The result is obtained on human-annotated reference summaries using a leave-one-out strategy~\cite{otani2019rethinking}, which measures the average consistency of human annotators. Although humans outperform all baseline models in most evaluation metrics on different tasks (except for BLEU@4), our proposed VTSUM-BLIP archives quite competitive results, especially on the V2T-SUM task. Fig.~\ref{fig:vis} visualizes some examples of generated video and text summaries, showing the effectiveness of TT and CA modules. Additionally, more visualization results are present in Fig.~\ref{fig:add_vis}.

\vspace{-5mm}
\subsection{Experimental Analysis}

\noindent \textbf{Method Comparisons on Existing Benchmarks}. To further evaluate the effectiveness of our proposed model, we conduct experiments on two well-known video summarization datasets, \ie, TVSum and SumMe. The results in TABLE~\ref{tab:v2vsum_table} show that the BLIP-VSUM (Base) achieves competitive results against several strong baselines. Moreover, our proposed mechanisms of Temporal Transformer and Context Aggregation can further improve video summarization performance. We also verify the ability of video-to-text summarization of our model on ActivityNet Captions. As we can see from TABLE~\ref{tab:v2tsum_table}, our proposed model outperforms all the strong baseline models by a large margin in multiple evaluation metrics (2.9 in BLEU@4, 1.0 in METEOR, 7.4 in ROUGE-L, and 19.0 in CIDEr).

\noindent \textbf{Human Evaluation}.
We conducted a human evaluation of video/text summaries on 50 random samples, assessed by workers for quality and consistency across video and text summaries, scoring 1-5 (5 best). We report the average score in TABLE~\ref{tab:he}. We can conclude from the table that our proposed model can generate more fluent and accurate text summaries for long videos. The proposed temporal Transformer and Context Aggregation can help generate accurate and consistent video summaries. Following \cite{wang2019controllable}, we compute the Kappa coefficient of different workers, and the value is 0.49 $\in$ (0.41, 0.60),
which means that the consistency is moderate.

\begin{table}[t]
\small
\centering
\caption{Human evaluation of the baseline models on the VideoXum dataset.}
\vspace{-2mm}
\begin{adjustbox}{width=0.99\linewidth}\setlength\tabcolsep{2pt}
\begin{tabular}{lccccc}
\toprule[1.2pt]
\multicolumn{1}{c}{\multirow{2}{*}{Method}} & \multicolumn{1}{c}{\bf V2V-SUM} & \multicolumn{2}{c}{\bf V2T-SUM} & \multicolumn{1}{c}{\bf Cross-Modal}\\
 \cmidrule(r){2-2} \cmidrule(r){3-4} \cmidrule(r){5-5}
\multicolumn{1}{c}{} & Accuracy  & Accuracy  & Fluency  & Consistency \\
\midrule[1.2pt]
VTSUM-BLIP (Base)&3.1&4.1&4.3&3.2\\
+ Temporal Transformer&3.5&4.2&4.2&3.2\\
+ Context Aggregation&3.2&4.1&4.3&3.1\\
+ TT + CA &\bf{3.8}&\bf{4.4}&\bf{4.4}&\bf{3.4}\\
\bottomrule[1.2pt]
\end{tabular}
\end{adjustbox}
\label{tab:he}
\end{table}

\begin{table}[t]
\centering
\caption{Results of cross-modality similarity under different semantic changes.}
\vspace{-2mm}
\begin{adjustbox}{width=0.99\linewidth}\setlength\tabcolsep{10pt}
\begin{tabular}{lcc}
\toprule[1.2pt]
\multirow{2}{*}{Method} & \multicolumn{2}{c}{\bf Cross-Modal Similarity (Cosine Similarity)}   \\ \cmidrule{2-3} 
                        & Positive pairs & Negative pairs \\ \midrule[1.2pt]
CLIPScore~\cite{hessel2021clipscore}                  & 14.5           &      0.3        \\
VT-CLIPScore                 & 38.0          & 0.2             \\ \bottomrule[1.2pt]
\end{tabular}
\label{tab:clipscore}
\end{adjustbox}
\vspace{-3mm}
\end{table}

\noindent \textbf{Comparison between CLIPScore and VT-CLIPScore}.
Although we can apply a pretrained CLIP model without any adaptation on our dataset to evaluate the semantic consistency of the video and text summaries, the similarity score may be insensitive to the semantic change of generated cross-modal summaries.
In TABLE~\ref{tab:clipscore}, we compare the vanilla CLIPScore~\cite{hessel2021clipscore} and the finetuned VT-CLIPScore to measure the similarity of different video and text summarization pairs.
The positive pairs refer to the paired video and text summaries. The negative pair includes unpaired video and text summaries.
From the results, we can observe that CLIPScore provides a solid foundation for the finetuned VT-CLIPScore.
Moreover, adapting the vanilla CLIP model to our proposed task is necessary since there is a domain gap between the CLIP pretraining data and our VideoXum data. The finetuned CLIP model on our data makes the evaluation score more reliable.
TABLE~\ref{tab:clipscore} shows that finetuning on VideoXum makes the similarity scores more reflective or informative in measuring the semantic consistency of cross-modal summaries.

\begin{figure*}[t]
    \centering
    \begin{minipage}{0.31\textwidth}
        \includegraphics[width=\textwidth]{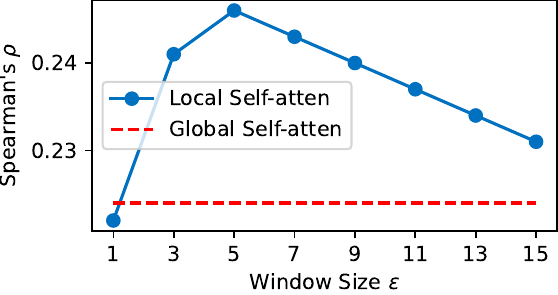}
        \caption{Impact of local window size $\varepsilon$ for Context Aggregation (CA) module.}
        \label{fig:win_size}
    \end{minipage}
    \hfill
    \begin{minipage}{0.65\textwidth}
        \includegraphics[width=0.29\textwidth]{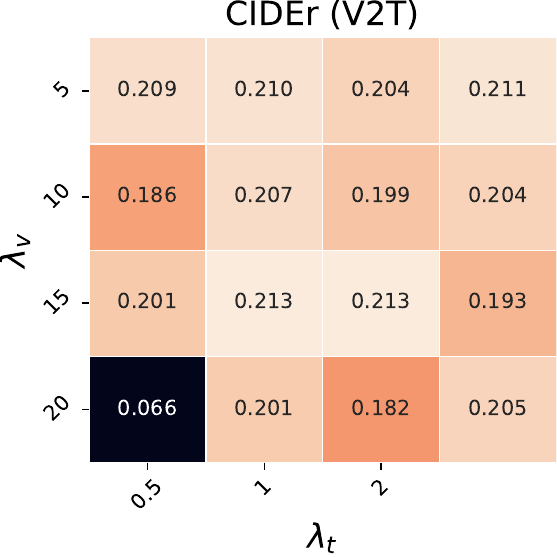}
        \hfill
        \includegraphics[width=0.29\textwidth]{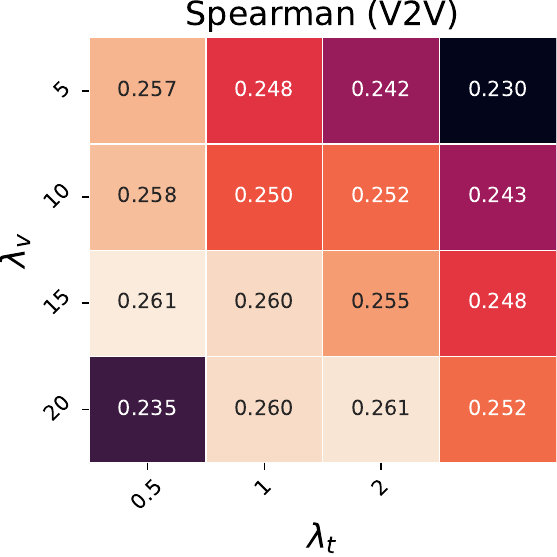}
        \hfill
        \includegraphics[width=0.29\textwidth]{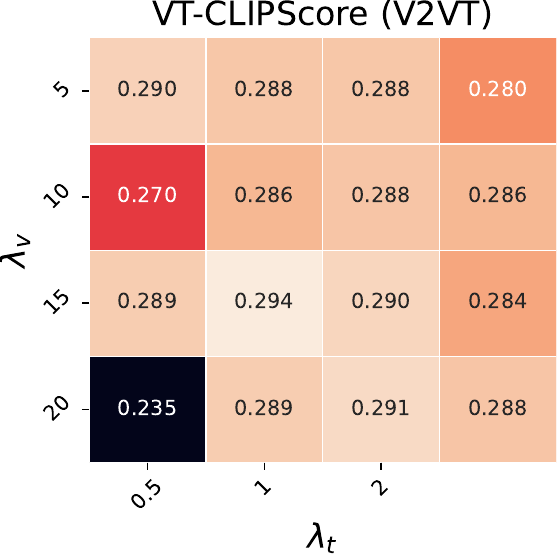}
        \vspace{-1mm}
        \caption{Impact of multi-task weights $\lambda_v$ and $\lambda_t$.} \label{fig:multitask_weight}
    \end{minipage}
\end{figure*}

\begin{table*}[t]
\centering
\vspace{-1mm}
\caption{Impact of Temporal Transformer (TT) Layers $N_\text{tem}$ on the VideoXum dataset \textit{val} set. The F1 score, BLEU@4, METEOR, ROUGE-L, CIDEr, and VT-CLIPScore are shown in \%.}
\vspace{-1.5mm}
\small
\begin{adjustbox}{width=0.99\textwidth}\setlength\tabcolsep{12pt}
\begin{tabular}{lcccccccc}
\toprule[1.2pt]
\multirow{2}{*}{Method} & \multicolumn{3}{c}{\textbf{V2V-SUM}} & \multicolumn{4}{c}{\textbf{V2T-SUM}} & \textbf{V2VT-SUM}   \\ \cmidrule(r){2-4} \cmidrule(r){5-8} \cmidrule(r){9-9}
                        & F1 score     & Kendall     & Spearman    & BLEU@4  & METEOR  & ROUGE-L  & CIDEr & VT-CLIPScore   \\ \midrule[1.2pt]
VTSUM-BLIP (Base)          & 21.9 &	0.157 &	0.208          &5.2 &	11.3 &	24.3 & 	18.3 & 28.4                \\
$\ \ $+ 1-layer TT  & 22.6       & 0.176          & 0.232          & 5.2      & 11.5      & 24.3       & 20.2    & 28.9                \\
$\ \ $+ 2-layer TT      & 22.7 & 0.176 & 0.232 & 5.2 &	11.4 &	   24.3       & 20.2  &	28.9 \\
$\ \ $+ 3-layer TT       & 22.7 & 0.174 & 0.230 & 5.1 &	11.4 &	   24.3       & 20.5  &	29.0 \\
\bottomrule[1.2pt]
\end{tabular}
\end{adjustbox}
\label{tab:tt_layer}
\vspace{-3mm}
\end{table*}

\vspace{-2mm}
\subsection{Extended Discussion}
{
\noindent\textbf{Complexity analysis}.
In terms of computational complexity, the V2V-Sum, V2T-Sum, and V2VT-Sum models require 0.06, 1.15, and 1.18 GPU hours for training on an A100 GPU, respectively. Regarding the model complexity, they have parameter sizes of 435.6 MB, 564.8 MB, and 567.1 MB.
}

{
\noindent\textbf{Impact of multi-task weights in Eq.(\ref{eq:loss})}.
To determine the impact of multi-task weights (\ie, $\lambda_v$ and $\lambda_t$), we perform an ablation study on the $\lambda_v$ and $\lambda_t$ using the VideoXum \textit{val} set. Fig.~\ref{fig:multitask_weight} shows that the peak point appears when $\lambda_v = 15.0$ and $\lambda_t = 1.0$ as mentioned in Section~\ref{sec:exp_setup}.
}

\noindent\textbf{Impact of Local Window Size $\varepsilon$}.
For the V2V-SUM task, the local window size $\varepsilon$ controls the context range of local self-attention. For $\varepsilon = 1$, the local self-attention module degrades to a multilayer perceptron (MLP). As $\varepsilon$ increases to $T$ ($>1$), the local self-attention module upgrades to a global/regular self-attention module.
Fig.~\ref{fig:win_size} shows the optimal performance occurs at $\varepsilon=5$; below this, limited context hinders performance, while above it, excess context introduces irrelevant frames, reducing efficacy. Therefore, the performance of the V2V-SUM task is improved by carefully selecting appropriate local context information.

\noindent\textbf{Impact of Temporal Transformer Layers}. We conduct an ablation study on the number of Temporal Transformer layers $N_\text{tem}$ using VideoXum \textit{val} set. TABLE~\ref{tab:tt_layer} indicates that altering $N_\text{tem}$ does not significantly affect the performance for all three tasks. Therefore, we set $N_\text{tem}$ to 1.

\noindent\textbf{Analysis for Human Performance on V2T-SUM}.
Human performance on our proposed three tasks can be regarded as an upper bound of each task.
TABLE~\ref{tab:videoxum_table} presents human performance outperforms our proposed model by a large margin on V2V-SUM and V2VT-SUM. However, on V2T-SUM, human performance does not exhibit a significant advantage over our model (especially on BLEU@4).
To better understand this phenomenon, we examine human-annotated captions and their corresponding references, where ``Human'' indicates the human predictions on VideoXum \textit{test} set and ``Reference'' denotes the corresponding ground truth. Both ``Human'' and ``Reference'' are human-annotated text summaries from ActivityNet Captions~\cite{krishna2017dense} validation sets. In particular, we present a representative example below:

\vspace{-2mm}

{\scriptsize
\begin{verbatim}
============================================================
*Human*: Two children stand in front of a mat. They throw
something onto the mat. They take turns jumping across the
mat. They pick up the item they threw on it.
============================================================
*Reference*: Two young children are standing in line indoors
in what looks like a living room. The little girl is stand-
ing closest to the hopscotch mat and she throws her toy onto
the mat and then begins jumping until she meets the end of
the mat then turns around and heads back to the point she
started and her turn is over. The little boy goes next, and
he throws the toy onto the mat and begins jumping to the end
of the mat, then turns around and jumps back towards his
starting point.The little girl steps in front of the boy and
gets into motion to start another turn on the hopscotch mat.
============================================================
\end{verbatim}
}

\vspace{-2mm}

\noindent Both captions describe two children playfully interacting on a mat, but ``Reference'' provides a more vivid and detailed picture of the scene. The comparison shows that summarizing a long video is inherently subjective, leading to varying text descriptions of the same content among different individuals. Therefore, it explains why human performance does not exhibit a significant advantage
over VTSUM-BLIP model.

\section{Conclusion}
In this study, we first propose a new video and text summarization task along with an enriched dataset VideoXum. In this task, we jointly consider the generic video summarization task and video-to-text summarization task. Furthermore, we propose a strong baseline framework VTSUM-BLIP to address the challenges for our proposed task. The empirical results show that our proposed framework achieves promising performance on VideoXum. In addition, we adapt CLIPScore on the VideoXum benchmark and introduce a new metric VT-CLIPScore to evaluate cross-modal video summarization semantic consistency, which shows high consistency with human evaluation on multiple experimental results.
For future studies, there are several promising directions on this benchmark.
There is plenty of room to explore the strategy of associating V2V and V2T summarization tasks for better performance and efficiency.
The proposed VideoXum dataset provides a foundation that could be significantly expanded through GPT-4~\cite{openai2023gpt4}, for generating video instruction-following data and thereby promoting the development of a general-purpose video assistant. 
For the evaluation metric, the adapted CLIP model for measuring video-text similarity is a practical compromise for the lack of large video-text pretrained models. It also suggests the need for a more reliable metric for video-text coherence measurement.
In addition, more advanced visual/video encoders and large language models (LLMs)~\cite{openai2023gpt4,touvron2023llama} could be integrated into the proposed framework to benefit the results of cross-modal summarization.
\clearpage
\clearpage

{\small
\bibliographystyle{IEEEtran}
\bibliography{egbib}
}

\begin{IEEEbiography}[{\includegraphics[width=1in,height=1.25in,clip,keepaspectratio]{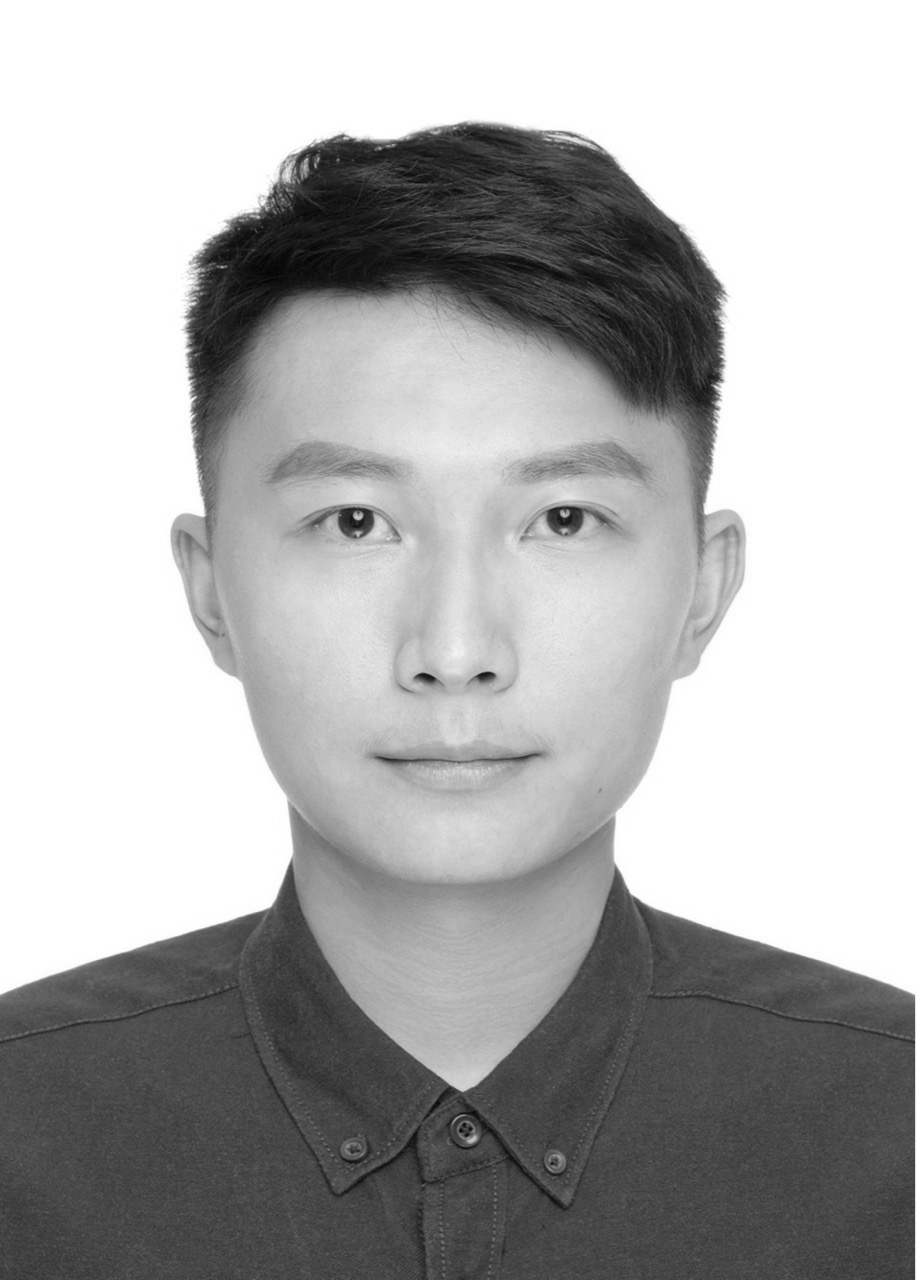}}]{Jingyang Lin} 
received the BE and MSc degree from Sun
Yat-sen University (SYSU), Guangzhou, China, in
2019 and 2022. He is currently a
PhD student in Computer Science at the University of Rochester. His research interests include
vision-and-language, video analysis, multi-task learning, and self-supervised learning.
\end{IEEEbiography}

\begin{IEEEbiography}[{\includegraphics[width=1in,height=1in,clip,keepaspectratio]{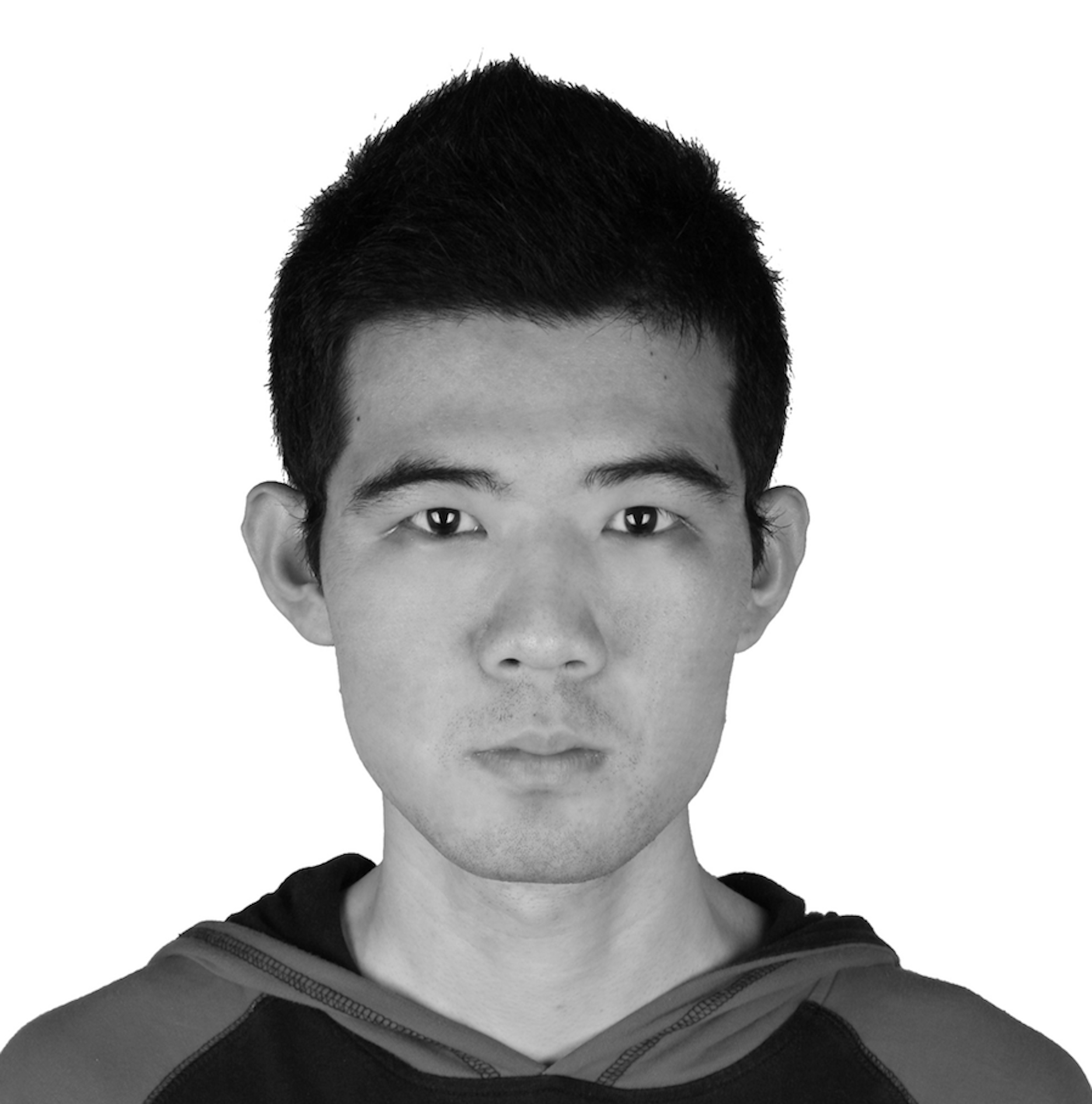}}]{Hang Hua}
 is a Ph.D. student in computer science at the University of Rochester. He is currently a member of the VIStA group advised by Professor Jiebo Luo. Hang got his master's degree from Peking University and his bachelor's degree from the South China University of Technology. His research interests are Vision and Language, machine learning, and social media analysis. 
\end{IEEEbiography}

\begin{IEEEbiography}[{\includegraphics[width=1in,height=1.25in,clip,keepaspectratio]{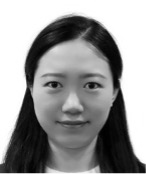}}]{Ming Chen}
is a Senior Staff Engineer at OPPO US Research Center.  She got a Ph.D. (2018) in Mechanical and Aerospace Engineering from University of California Davis.  After that she joined OPPO US Research Center.  Her research interests include video understanding, video summarization/highlight detection and visual semantic representation learning.
\end{IEEEbiography}

\begin{IEEEbiography}[{\includegraphics[width=1in,height=1.25in,clip,keepaspectratio]{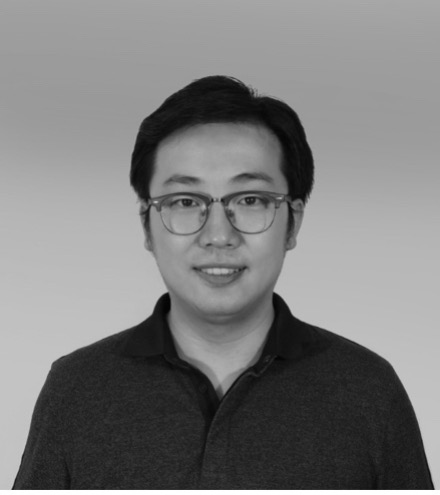}}]{Yikang Li}
is a Senior Research Engineer at the OPPO US Research Center. His research interests include video understanding, multi-modal foundational model learning, and visual semantic representation learning. Yikang received Ph.D. (2020) in Electrical Engineering from Arizona State University, and during his Ph.D., he investigated different video semantic representations via deep learning techniques.
\end{IEEEbiography}

\begin{IEEEbiography}[{\includegraphics[width=1in,height=1.25in,clip,keepaspectratio]{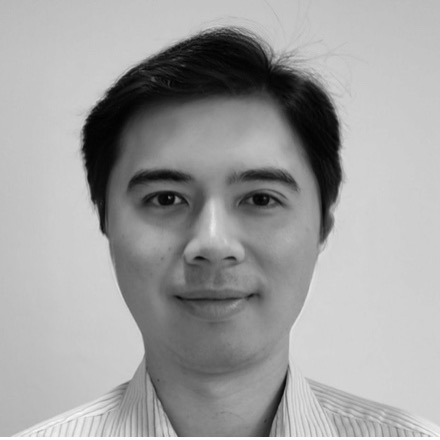}}]{Jenhao Hsiao}
received the PhD degree in Electrical Engineering \& Computer Science from National Taiwan University. He is currently a principle research scientist with OPPO US Research Center. Before joining OPPO, he held research positions in Yahoo! Research and IBM Research. His main research interests include machine learning, deep learning, computer vision, vision-language model (VLMs), and other related problems in AI.
\end{IEEEbiography}

\begin{IEEEbiography}[{\includegraphics[width=1in,height=1.25in,clip,keepaspectratio]{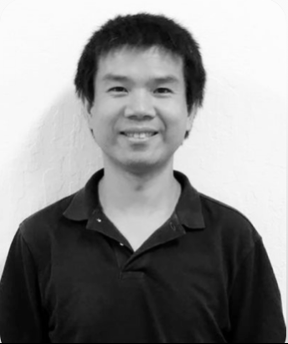}}]{Chiuman Ho}
is the Senior Director of AI at OPPO US Research Center. He got a PhD in theoretical physics from University of Pittsburgh and did a postdoc at UC Berkeley. Before leaving the academia, he was a Research Assistant Professor at Michigan State University. Prior to joining OPPO, he was a Senior Staff Data Scientist at Huawei US R\&D. Chiuman is highly interested in AI research and its commercialization. He also helps to develop AI strategies for OPPO, and leads a team to apply AI to improve user experience.
\end{IEEEbiography}

\begin{IEEEbiography}[{\includegraphics[width=1in,height=1.25in,clip,keepaspectratio]{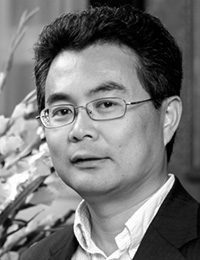}}]{Jiebo Luo (Fellow, IEEE)}
is the Albert Arendt Hopeman Professor of Engineering and Professor of Computer Science at the University of Rochester which he joined in 2011 after a prolific career of fifteen years at Kodak Research Laboratories. He has authored nearly 600 technical papers and holds over 90 U.S. patents. His research interests include computer vision, NLP, machine learning, data mining, computational social science, and digital health. He has been involved in numerous technical conferences, including serving as program co-chair of ACM Multimedia 2010, IEEE CVPR 2012, ACM ICMR 2016, and IEEE ICIP 2017, as well as general co-chair of ACM Multimedia 2018 and IEEE ICME 2024. He has served on the editorial boards of the IEEE Transactions on Pattern Analysis and Machine Intelligence (TPAMI), IEEE Transactions on Multimedia (TMM), IEEE Transactions on Circuits and Systems for Video Technology (TCSVT), IEEE Transactions on Big Data (TBD), ACM Transactions on Intelligent Systems and Technology (TIST), Pattern Recognition, Knowledge and Information Systems (KAIS), Machine Vision and Applications, and Intelligent Medicine. He was the Editor-in-Chief of the IEEE Transactions on Multimedia (2020-2022). Professor Luo is also a Fellow of NAI, ACM, AAAI, SPIE, and IAPR.
\end{IEEEbiography}

\end{document}